%% file: manuscript-JT.tex
\documentclass[conference]{IEEEtran}
\IEEEoverridecommandlockouts
\usepackage{cite}
\usepackage{amsmath,amssymb,amsfonts}
\usepackage{algorithmic}
\usepackage{graphicx}
\usepackage{textcomp}
\usepackage{array}
\usepackage{booktabs}
\usepackage{multirow}
\usepackage{threeparttable}
\usepackage{subcaption}
\usepackage{xcolor}

\usepackage{tabularx}
\usepackage{rotating}
\usepackage{caption}
\usepackage{wrapfig}
\usepackage[normalem]{ulem}

\usepackage{tikz}
\usepackage{textcomp}
\usepackage{hyperref}
\usepackage{lipsum}

\newcommand\copyrighttext{%
  \footnotesize \textcopyright 2024 IEEE. Personal use of this material is permitted.
  Permission from IEEE must be obtained for all other uses, in any current or future
  media, including reprinting/republishing this material for advertising or promotional
  purposes, creating new collective works, for resale or redistribution to servers or
  lists, or reuse of any copyrighted component of this work in other works.
  DOI: 10.1109/CogMI62246.2024.00035}
\newcommand\copyrightnotice{%
\begin{tikzpicture}[remember picture,overlay]
\node[anchor=south,yshift=10pt] at (current page.south) {\fbox{\parbox{\dimexpr\textwidth-\fboxsep-\fboxrule\relax}{\copyrighttext}}};
\end{tikzpicture}%
}

\def\BibTeX{{\rm B\kern-.05em{\sc i\kern-.025em b}\kern-.08em
    T\kern-.1667em\lower.7ex\hbox{E}\kern-.125emX}}
    
\begin{document}

\title{Decoding Linguistic Nuances in Mental Health Text Classification Using Expressive Narrative Stories}

\author{
    \IEEEauthorblockN{Jinwen Tang\IEEEauthorrefmark{1}, Qiming Guo\IEEEauthorrefmark{2}, Yunxin Zhao\IEEEauthorrefmark{1}, and Yi Shang\IEEEauthorrefmark{1}}
    
    \IEEEauthorblockA{\IEEEauthorrefmark{1}\textit{Electrical Engineering and Computer Science Department} \\
    \textit{University of Missouri}\\
    Columbia, Missouri, USA \\
    \{jt4cc, zhaoy, shangy\}@umsystem.edu}

    \IEEEauthorblockA{\IEEEauthorrefmark{2}\textit{Department of Computer Science} \\
    \textit{Texas A\&M University-Corpus Christi}\\
    Corpus Christi, Texas, USA \\
    qguo2@islander.tamucc.edu}
}

\maketitle

\copyrightnotice

\begin{abstract}
Recent advancements in Natural Language Processing (NLP) have spurred significant interest in analyzing social media text data for identifying linguistic features indicative of mental health issues. However, the domain of Expressive Narrative Stories (ENS)—deeply personal and emotionally charged narratives that offer rich psychological insights—remains underexplored. This study bridges this gap by utilizing a dataset sourced from Reddit, focusing on ENS from individuals with and without self-declared depression.
Our research evaluates the utility of advanced language models, BERT and MentalBERT, against traditional models such as SVM, Naive Bayes, and Logistic Regression. We find that traditional models are notably sensitive to the absence of explicit topic-related words, which could risk their potential to extend applications to emotional expressive narratives that lack clear mental health terminology. Despite MentalBERT’s design to better handle psychiatric contexts, it demonstrated a dependency on specific topic words for classification accuracy, raising concerns about its application in scenarios where explicit mental health terms are sparse (P-value \textless{} 0.05). In contrast, BERT(128) exhibited minimal sensitivity to the absence of topic words in ENS, suggesting its superior capability to understand deeper linguistic features, making it more effective for real-world applications that require nuanced text analysis.
Both BERT and MentalBERT excel at recognizing linguistic nuances and maintaining classification accuracy even when narrative order is disrupted—a crucial capability in mental health narratives. This resilience is statistically significant, with sentence shuffling showing substantial impacts on model performance (P-value \textless{} 0.05), especially evident in ENS comparisons between individuals with and without mental health declarations.
These findings underscore the importance of exploring ENS for deeper insights into mental health-related narratives, advocating for a nuanced approach to mental health text analysis that moves beyond mere keyword detection. This study paves the way for more sophisticated, context-aware analyses in NLP applications, aiming to enhance the understanding of linguistic patterns associated with mental health conditions.
\end{abstract}


\section{Introduction}
Mental health disorders represent a significant public health challenge, impacting over 20\% of adults in the United States each year. Conditions such as depression, bipolar disorder, and anxiety disorders profoundly affect individuals' physical and social well-being \cite{RN679}. This situation is exacerbated by the difficulty in providing effective assistance to those in need, a challenge partly rooted in the stigma surrounding mental health.

The digital age, particularly social media platforms like Twitter, Reddit, and Weibo, has provided novel avenues for mental health research. These platforms reveal insights into individuals' mental states through their natural language expressions, yielding a vast and diverse data source. This study focuses on Expressive Narrative Stories (ENS), a rich, yet under-explored domain within this context. Originating from Pennebaker's concept of Expressive Writing and Narrative Stories \cite{pennebaker1986confronting}, ENS offer profound psychological insights. Documented benefits of expressive writing, a practice involving writing about one's feelings and emotions related to a personal event or interaction, include improved physical health markers like reduced blood pressure \cite{davidson2002expressive} and enhanced immune functioning \cite{pennebaker1988disclosure, rivkin2006effects}, with narrative stories being crucial for processing experiences and constructing coherent self-identities \cite{mcadams2013narrative}.

Despite the richness of ENS and their known benefits, their potential in identifying mental health issues, particularly on social media, is not fully realized. Our study seeks to explore the linguistic nuances of ENS among individuals with and without mental health disorders. We hypothesize that these nuances will offer novel insights into mental health and natural language analysis.

This study will analyze Reddit text posts, contrasting Expressive Narrative Stories (ENS) from individuals with self-declared mental health disorders against those from ostensibly healthy users. Our objectives include: 1) fine-tuning (or training) and evaluating various machine learning models, such as Bidirectional Encoder Representations from Transformers (BERT), MentalBERT, Naive Bayes (NB), Logistic Regression (LR), and Support Vector Machine (SVM). These models are intended to classify text posts based on whether they come from individuals with or without self-declared mental health disorders, helping to identify linguistic features unique to these groups; 2) assessing the significance of topic words in distinguishing depression-related ENS; 3) examining the impact of sentence structure on narrative coherence in ENS; and 4) broadening our analysis to encompass multiple mental health disorders to uncover common linguistic features.

Key contributions of our research to the fields of mental health analysis and natural language processing are as follows:
\begin{itemize}
\item {\bfseries Focused Analysis on ENS:} By concentrating on ENS within social media, our research fills a significant gap, shedding light on the nuanced ways narrative storytelling intertwines with mental health.
\item {\bfseries In-depth Linguistic Exploration:} Our comprehensive examination of linguistic features such as topic words and sentence order in ENS establishes a foundation for subsequent studies, underscoring the critical role of narrative context in understanding mental health.
\item {\bfseries Model Comparative Evaluation:} Our detailed comparison of diverse machine learning models elucidates their strengths and weaknesses in mental health narrative text analysis, guiding future technological developments.
\end{itemize}

The insights derived from this study are poised to enhance mental health interventions on social media platforms, where individuals frequently articulate and share their experiences. The remainder of this paper details related work, describes our data and methodology, presents our findings, and concludes with a discussion of their broader implications.

\section{related work}

Recent advancements in machine learning and NLP have considerably influenced mental health research, with social media data emerging as a key resource for understanding mental health states. The natural language expressions on platforms like Twitter, Reddit, and Weibo provide invaluable data for linguistic analysis \cite{cavazos2016content, pennebaker2003psychological, hart2001redeveloping}.

The expansion of social media platforms like Twitter, Reddit, and Weibo has been a boon for mental health research. These platforms provide a unique window into the mental states of individuals, especially those reluctant to seek professional help \cite{andrews2001shortfall, regier1993facto}. The natural language expressions found on these platforms, ranging from everyday updates to personal stories, offer researchers a trove of data for linguistic analysis \cite{cavazos2016content, pennebaker2003psychological, hart2001redeveloping}.

Studies have increasingly employed sophisticated machine learning models and NLP techniques to analyze social media data. For instance, \cite{wang2013depression} employed sentiment analysis to identify depression-related posts on mini-blogs. \cite{yazdavar2017semi} extracted symptom-related features from Twitter to detect depression. Meanwhile, \cite{yates2017depression} and \cite{inamdar2023machine} successfully identified users with suicidal tendencies and stress on Reddit through machine learning models including CNN, BERT, and BoW. Additionally, platforms such as Sina Weibo \cite{wang2020multitask}, Facebook \cite{katchapakirin2018facebook}, and other online communities such as  OurDataHelps.org \cite{coppersmith2018natural}, TeenHelp.org \cite{franz2020using}, etc., have been leveraged to support mental health studies.  

While these computational methods have shown promise, there is a growing need for knowledge-based analysis to complement these approaches. Studies incorporating such analyses aim to enhance the interpretability of results, linking computational findings with psychological theories and clinical insights. For example, \cite{sampath2022data} were able to detect depression severity levels based on criteria assigned to their self-collected Reddit dataset. This integration of computational methods with clinically relevant criteria exemplifies the potential of combining machine learning with domain-specific knowledge. Similarly, \cite{li2021incorporating} incorporated predicted events and personality traits, extracted from microblogs, into their analysis, achieving a positive deviation from their baseline results. \cite{chaurasia2021predicting} used a BERT model to analyze offensive slang in social media posts, finding correlations with poor mental health outcomes.

Despite these advancements, there is a notable gap in the specific analysis of Expressive Narrative Stories (ENS) in social media. ENS provide a rich, narrative-driven context that may reveal deeper insights into the mental states of individuals, a potential yet to be fully explored in the realm of mental health research on social media.

\section{Research Approach}
The focus of many research studies on optimizing model accuracy without considering the relevance of the data often leads to misleading results, particularly in the field of mental health detection. Generic data such as news or jokes, which lack personal expression, are ill-suited for detecting mental health issues. Our approach emphasizes narratives that contain personal and emotional experiences, which are more indicative of an individual's mental state. By distinguishing between general narratives and those with emotional expressive content, our methodology aims to uncover nuanced linguistic features critical for accurate mental health assessment. This focus not only addresses the shortcomings of previous studies but also enhances the potential for real-world application, particularly in pre-screening scenarios where individuals may not explicitly discuss their mental health status. To effectively address these issues, our study employs a three-phase approach to deeply analyze linguistic features relevant to mental health assessment.

In this study, we followed a three-phase approach to examine linguistic features related to Expressive Narrative Stories (ENS) and other common narratives. After initially collecting ENS related to depression (Depression), other ENS (ENS), and general narratives (GNS) from Reddit, we proceeded as follows:

\begin{enumerate}
\item {\bfseries Classifier Implementation and Evaluation:} We implemented and evaluated several machine learning models, including BERT, MentalBERT, SVM, Naive Bayes, and Logistic Regression, to identify the most effective classifiers based on our datasets. The primary goal of these classifiers is to accurately categorize text posts into groups: those authored by individuals with self-declared mental health disorders and those by ostensibly healthy users. This phase focused on the foundational setup of our analytical tools and conducted preliminary analyses to determine which models perform best at distinguishing between these two categories based on linguistic features and patterns inherent in the text. This step is crucial for enabling further in-depth analysis of language use and mental health indicators within the Reddit posts.
\item {\bfseries Impact of Topic Words:} We assessed the impact of topic words on different datasets by examining the performance of classifiers across various word manipulations. Topic words are key terms or phrases that are highly relevant or frequently used in specific contexts or subjects discussed within the narratives. This phase involved detailed analysis to understand how these specific words influenced the classification results, thereby shedding light on the sensitivity of our models to content-specific elements within the narratives.
\item {\bfseries Logical Connection Analyses:} We investigated the logical connections between sentences, defined as the coherent and meaningful relationships that link sentences within a narrative. Our focus was on evaluating the performance of both BERT and MentalBERT classifiers across different sentence manipulation scenarios. This analysis included within-post shuffling and cross-post shuffling to simulate the cognitive disruptions typically observed in individuals with mental health issues. Additionally, we explored the generalization of our findings to other mental health-related expressive narratives, assessing how these classifiers perform under conditions of narrative disorganization.
\end{enumerate}

\subsection{Data Collection and Preprocessing}

To prevent data leakage—a common issue where data intended for validation or testing inadvertently influences the training process—we meticulously conducted data collection in batches. This method ensures that each batch is independent and used only in the appropriate phase of model training, validation, or testing, thereby preserving the integrity of our evaluation metrics.

Posts we collected from Reddit between late 2022 and early 2023 were utilized to create three distinct datasets for training and validation. Posts gathered from the r/AnxietyDepression subreddit were identified as target posts and assigned a label of 1. Conversely, all other posts were considered control posts and labeled as 0. To further enhance the study's robustness, two additional comparison datasets were established: one comprising other expressive narrative story posts (ENS) collected from subreddits such as r/IamA, r/relationship, and r/AmItheAsshole, and another consisting of general narrative story posts (GNS) sourced from subreddits like r/cooking, r/healthyfood, and r/eathealthy.

Before the training process, we applied several crucial preprocessing steps to the collected posts to ensure data quality and relevance. First, all text was converted to lowercase to maintain uniformity across the data. Next, we removed posts with fewer than 10 words and those containing redirected URL links (e.g., http) to eliminate advertisements and irrelevant content. Additionally, certain Unicode-related quotation marks, such as \textbackslash{}u2019 in "I\textbackslash{}u2019m" which indicates the single quotation mark ' ', were removed to standardize text formatting. We also removed all promotional posts and posts that did not contain first-person pronouns from the ENS dataset, which includes both mental health-related and other common narratives. Finally, manual checks were performed to ensure that only narrative posts were retained, aligning with the study's focus on expressive narratives. An example of the raw text after preprocessing is presented in Table I.
\input{table_I}

Three distinct datasets were created to train and validate models for this study as follows:
\begin{enumerate}
\item GNS-Depression dataset: This dataset consists of 1,650 general narrative story posts from subreddits and 1,076 posts from the r/AnxietyDepression subreddit.
\item ENS-Depression dataset: This dataset includes 1,110 posts from subreddits featuring common expressive narrative stories and 1,076 posts from the r/AnxietyDepression subreddit.
\item Mix-Depression dataset: This dataset combines all posts from the previous two datasets. The posts from the r/AnxietyDepression subreddit remain the same across all three datasets.
\end{enumerate}

The descriptive statistics for these datasets are presented in Table II, which includes information about the label ratios, indicating the proportion of control and target posts in each dataset. The distribution of word counts for all datasets is positively skewed. To ensure randomness and maintain label distribution, all three datasets were initially shuffled randomly. Subsequently, each dataset was split into a training set and a validation set using a split ratio of 7:3. Importantly, the distribution of target labels was maintained at the same ratio (as shown in Table II) for both training and validation sets within each dataset.

\input{table_II}

To create testing sets, we utilized newly collected common Expressive Narrative Stories (ENS), depression-related ENS, and General Narrative Stories (GNS) posts. Additionally, we gathered posts related to other mental health disorders for further generalization analyses, including posts on anxiety and bipolar disorder collected directly from the r/Anxiety and r/Bipolar subreddits, respectively. Suicide-related data was extracted from the SuicideWatch subreddit via a Kaggle dataset (https://www.kaggle.com/general/256134) \cite{Rigoulet2021}. The same data preprocessing methods applied to the training data were also employed for these new sets, ensuring consistency across all phases of the study. Detailed information about these 12 new testing and generalization sets is provided in Table III. The ratio of labels for all testing sets is 1:1.

\input{table_III}

\subsection{Phase 1: Fine-tuning and Comparing Models}
In this phase, we fine-tuned two models: the BERT-base model, which was previously trained on a lowercased English corpus \cite{devlin2018bert}, and its domain-specific extension, MentalBERT \cite{ji2021mentalbert}. We aimed to compare the performance of these models with traditional machine learning classifiers including Naïve Bayes, Logistic Regression, and SVM.

\subsubsection{\bfseries{Methods}}

We fine-tuned both the BERT-base uncased English-only model and MentalBERT using identical configurations to ensure a fair comparison. Each model has 12 hidden layers, a hidden size of 768, and 12 attention heads. For text vectorization, we utilized their built-in tokenizers with subword tokenization (WordPiece) and retained stopwords to preserve the full contextual meaning of the texts. The models were adjusted using three new datasets. Key hyperparameters were uniformly set for both models to optimize performance: the training batch size was 32, balancing computational efficiency and model convergence. Training was conducted over 5 epochs to balance learning and prevent overfitting. The learning rate was set at 2e-5, chosen to optimize weight adjustment during training. The Sigmoid function served as the output activation, allowing the models to produce probabilistic predictions. Additionally, the Adam optimizer was employed for its effectiveness in optimizing deep learning models.

Input sequence lengths of 10, 64, 128, and 300 tokens were tested, with a chosen length of 128 tokens as it is slightly greater than the average length of the posts and captures about half of the posts in full length. The lengths of 10 and 300 tokens were used to examine the performance of the model when trained with partial and more extensive content of the post, respectively, to determine their influence on model performance.

To evaluate the effectiveness of our fine-tuned BERT and MentalBERT models, we compared their performance against other frequently used statistical classifiers. The same training and testing datasets were employed for Naïve Bayes, Logistic Regression, and SVM models, which utilized features including TF-IDF of up to 5000 most frequent terms and Unigram features.

This  comparison helps in understanding the relative strengths of deep learning models like BERT and MentalBERT against traditional statistical methods in processing complex narrative datasets.

\subsubsection{\bfseries{Results}}

\input{table_IV}
An initial objective was to evaluate the effectiveness of both BERT and MentalBERT models across varying input token lengths and to compare their performance with traditional statistical models. We aimed to assess their ability to distinguish between Depression-related Expressive Narrative Stories (ENS) and other narrative types, such as General Narrative Stories (GNS), setting the stage for further investigations into mental health-related text analysis.

As demonstrated in Table IV, our fine-tuned models show strong performance across three distinct datasets, highlighted by high accuracy rates. We assess model effectiveness using key metrics, including accuracy—the ratio of correctly predicted observations to total observations—and the F1 score. The F1 score, a harmonic mean of precision and recall, is particularly valuable in scenarios with class imbalances as it ensures that both false positives and false negatives are considered.

Significantly, extending the input token length from 10 to 128 tokens markedly improved performance for the ENS-Depression and Mix-Depression testing sets. However, increasing the token count beyond 128 did not yield further improvements and, in some instances, resulted in a slight decrease in accuracy. The optimal results were obtained with 128 input tokens, where both BERT and MentalBERT models achieved an accuracy and F1 score of 0.96 in the ENS-Depression and Mix-Depression testing sets. Conversely, for the GNS-Depression testing set, performance plateaued at 64 tokens, maintaining an accuracy and F1 score of 0.99. This indicates that additional tokens did not contribute to further improvements in this dataset.

Interestingly, the traditional statistical models, particularly SVM with TF-IDF (5000 terms) and both Naive Bayes configurations, showed competitive performance in the GNS-Depression testing set, matching the deep learning models with accuracies and F1 scores of 0.99. However, in the ENS-Depression and Mix-Depression testing sets, the BERT and MentalBERT models consistently outperformed these statistical models, highlighting their superior capability in more complex comparative analyses of different types of expressive narrative content.

Our findings suggest that the BERT(128) and MentalBERT(128) models, configured with a maximum input token length of 128, serve as effective classifiers for detecting nuances in mental health narratives. Additionally, the clear distinction captured by simpler models between Depression-related expressive narratives and GNS underscores their potential utility in certain analytical scenarios. However, for more nuanced distinctions among various ENS, deeper learning parameters may be necessary to achieve higher accuracy. These insights are invaluable for our ongoing and future analyses in mental state NLP.

\subsection{Phase 2: Topic Words Manipulations}
To develop a predictive tool effective even when users do not explicitly mention mental health topics or symptoms, this phase focused on the impact of topic words on classifier performance across different datasets. We investigated how variations in keyword presence influenced the classifiers' ability to distinguish between Depression-related Expressive Narrative Stories (Depression), other common ENS (ENS), and General Narrative Stories (GNS).

Using the WordCloud method, we generated a list of the most influential topic-related words from each training dataset. We created a word list that included ten key words for each training set, along with their variations. These words underwent two types of manipulations across all three testing sets (Table V):
\begin{enumerate}
\item {\bfseries Words Removing:} Removing all these words from all testing sets 
\item {\bfseries Words Replacing:} Replacing all these words to a neutral word “nothing” for all testing sets.  
\end{enumerate}
\input{table_V}

\subsubsection{\bfseries{Results}}
The results presented in Table VI provide a comprehensive view of how different models respond to word manipulations in their ability to classify various types of narrative content. This analysis is particularly crucial in understanding the models' robustness and their applicability in real-world pre-screening for mental health issues.
\input{table_VI}

Across all three testing sets, traditional shallow learning models and MentalBERT displayed clear sensitivity to the manipulation of topic words, as evidenced by significant results in the paired t-test (P-value \textless{} 0.05). The sensitivity of these models to word manipulation was reflected in the performance metrics, where even modest manipulations resulted in noticeable changes in accuracy and F1 scores. Specifically, in the Mix-Depression testing set, both word removal and replacement had a significant impact on the performance of most models, including MentalBERT, except for the word replacement process under the Naive Bayes (Unigram) model. For the ENS-Depression testing set, word removal was particularly impactful, except for the Naive Bayes (TF-IDF:5000) model, which showed a notable response to word replacement. In the GNS-Depression testing set, it was primarily the word replacement manipulation that influenced model performance, with the Logistic Regression (TF-IDF:5000) and SVM (Unigram) models also showing significant sensitivity to both types of manipulations.

Interestingly, despite MentalBERT's design to better handle psychiatric context-specific language, it still demonstrated a dependency on specific topic words for classification accuracy. This raises concerns about its practicality in real-world applications, where explicit mentions of mental health terms may be sparse or absent. Unlike MentalBERT, the BERT(128) model exhibited a distinct behavior; it showed minimal sensitivity to word manipulations in the ENS-Depression testing set. This suggests its superior capability to capture more nuanced linguistic features beyond the presence of specific keywords, making BERT(128) potentially more effective for applications that require a deeper understanding of context and subtleties in language use—attributes common in real-world scenarios.

These findings indicate that while MentalBERT and traditional shallow learning models are sensitive to the presence or absence of specific topic words, BERT(128) provides a more stable and insightful analysis across various emotional expressive narrative contexts. BERT(128)'s deep learning architecture enables it to discern nuanced linguistic patterns that are crucial for accurately identifying mental health issues from narratives that do not explicitly mention mental health, without relying heavily on specific terminology.

In summary, the varying performances across different word manipulations highlight the diverse capabilities and limitations of the models tested. While shallow learning models and MentalBERT rely significantly on specific word patterns, making them susceptible to manipulations, BERT(128) demonstrates a capacity to handle linguistic variations effectively. This underlines the importance of selecting a model that aligns with the specific requirements of mental health text analysis, especially in settings where conventional keyword-based approaches may fall short.

\subsection{Phase 3: Logical Connection Analyses}
The structural integrity of narratives provides significant insights into cognitive processes that are intricately linked to both psycholinguistics and cognitive science. Emotional states and mental health conditions profoundly influence these processes, often altering logical reasoning abilities. Jung et al. \cite{jung2014emotions} have established that emotional disturbances can significantly impair logical reasoning capabilities. Videbeck expands on this by noting that individuals suffering from severe anxiety may encounter profound disruptions in reasoning, manifesting as disorganized thoughts that directly impact the coherence and logical flow of their narratives \cite{videbeck2010psychiatric}. This phase of our research leverages NLP methods to delve into these phenomena by examining how the logical connections between sentences are influenced by sentence manipulations and assessing the performance implications on BERT and MentalBERT classifiers. Additionally, we explore the potential for these findings to be generalized to other mental health-related expressive narrative stories (ENS).

\subsubsection{\bfseries{Methods}}
\begin{figure}[t]
  \centering
  \includegraphics[width=\columnwidth]{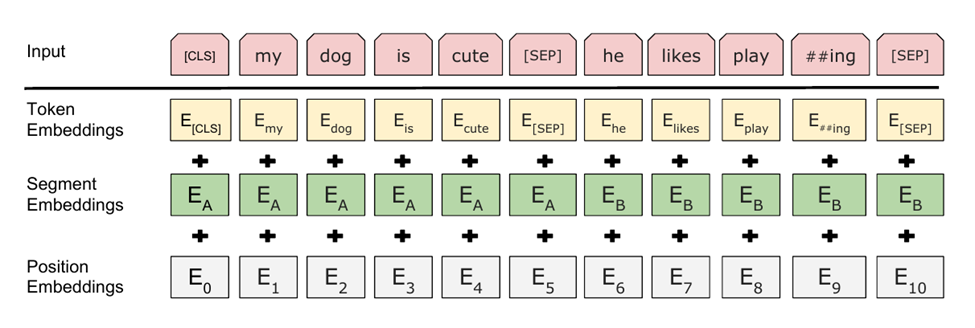} 
  \caption{Illustration of BERT Embeddings \cite{devlin2018bert}}
\end{figure}

Utilizing the advanced bidirectional capabilities of BERT and MentalBERT models, which are seminal in analyzing textual contexts, allows for a sophisticated examination of narrative coherence and logical connectivity. These models integrate token, segment, and position embeddings, facilitating a dynamic analysis by considering textual information from both directions simultaneously (Figure 1). This bidirectional approach is pivotal in detecting subtleties in narrative structures that may indicate cognitive disruptions typical of various mental health conditions.

To comprehensively test the impact of sentence ordering on narrative integrity, we implemented two randomization techniques across all 12 testing sets, as illustrated in Fig. 2:

\begin{itemize}
\item {\bfseries Within-Post Shuffling:} This method involves randomly shuffling sentences within each post to disrupt the inherent narrative flow. It is designed to test the models' capabilities to deduce context and sustain classification accuracy amid non-linear logical progression, thereby simulating the type of cognitive disruptions observed in individuals with mental health disorders such as severe anxiety.
\item {\bfseries Cross-Post Shuffling:} By shuffling sentences across posts that share the same label, this method introduces a layer of severe narrative disorganization. It evaluates the models' ability to still discern underlying themes or emotional tones that align with specific mental health conditions despite considerable structural disarray.
\end{itemize}

\begin{figure}
  \centering
  \includegraphics[width=\columnwidth]{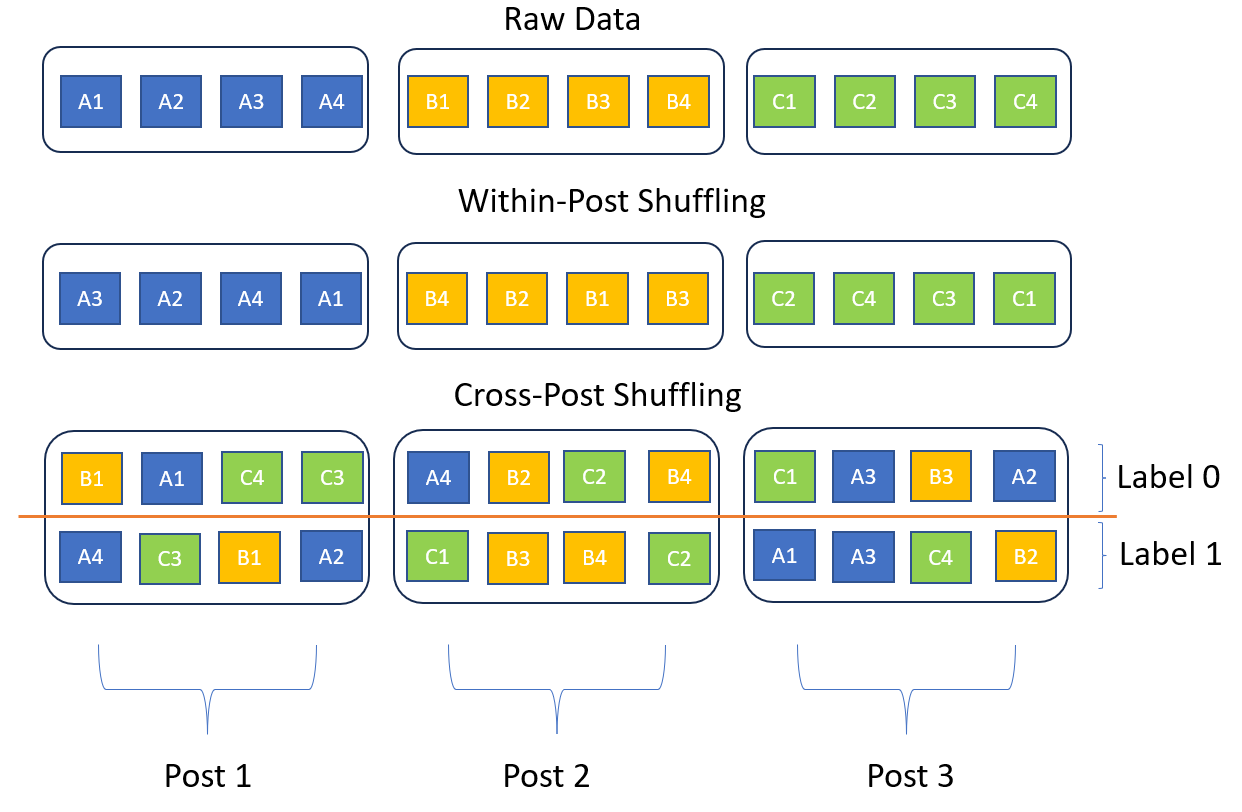} 
  \caption{Illustration of Two Sentence Shuffling Manipulations}
\end{figure}

We used BERT(128) and MentalBERT(128) over the 300 token variants due to their superior performance in handling complex semantic structures across all testing sets in Phase 1. Moreover, the positively skewed nature of our datasets suggests that a maximum input length of 300 tokens would introduce an excessive number of padding tokens, potentially diluting the meaningful content with irrelevant data and affecting the accuracy of model predictions.

This framework does not merely test the resilience of BERT and MentalBERT to variations in logical structuring but also aligns with clinical observations of significant deteriorations in narrative coherence within mental health disorders. By analyzing how these models manage such disruptions, we aim to illuminate their potential applications in real-world mental health assessments, where narrative inputs frequently exhibit logical inconsistencies or thematic disjunctions.

In addition, we synthesized data from three additional mental health issues—Anxiety, Bipolar, and Suicide—with three control datasets—GNS, ENS, and Mix—to construct nine additional testing sets. This strategic combination broadens the scope of our investigation, enhancing the generalizability of our findings across a more diverse spectrum of mental health conditions. The description of these 12 testing set was shown in Table III.

\subsubsection{\bfseries{Results}}
\input{table_VII}
Table VII presents the classification performance of the models and the results of paired t-tests between the raw data and Within-Post Shuffling, as well as Cross-Post Shuffling. It's evident that both sentence randomization processes had no discernible effect across all GNS-related mental health testing sets. However, for ENS-related testing sets, significant impacts were observed for both Within-Post Shuffling and Cross-Post Shuffling, with the Within-Post Shuffling showing an impact with a p-value of less than 0.05 and the Cross-Post Shuffling showing significant differences with a p-value of less than 0.01. The Mix-related testing set, which includes a mix of common ENS, appeared to be significantly influenced by the Cross-Post Shuffling but less so by the Within-Post Shuffling.

The BERT(128)-GNS model showed no significant changes in response to its corresponding dataset. Interestingly, the Cross-Post Shuffling appeared to slightly enhance the classification accuracy for this model. Both the BERT(128)-ENS and BERT(128)-Mix models exhibited notable sensitivity to the sentence randomization processes, with the Within-Post Shuffling causing a greater deviation in model performance.

These findings suggest that the logical connection between sentences is crucial in distinguishing ENS from narratives of individuals with and without depression, consistent across various mental health disorders. However, altering the sentence order in GNS-related testing sets did not significantly impact classification performance, aligning with results from Phase 2 where BERT(128) demonstrated minimal sensitivity to word manipulations in the ENS-Depression testing set. This underscores BERT(128)'s capability to capture more nuanced linguistic features beyond the presence of specific keywords, which is crucial for accurately identifying mental health issues in narratives that do not explicitly mention mental health concerns.

The contrast in model behavior between BERT(128) and MentalBERT also highlights the different sensitivities to linguistic manipulations. While MentalBERT showed dependency on specific topic words for classification accuracy, BERT(128) provided a more stable and insightful analysis across varied emotional expressive narrative contexts. This finding further underscores the importance of selecting an appropriate model for mental health text analysis, especially in scenarios where traditional keyword-based approaches might be insufficient.

In summary, the variation in performance across different sentence and word manipulations underscores the diverse capabilities and limitations of the models tested. While shallow learning models and MentalBERT remain sensitive to specific word patterns, BERT(128) demonstrates a robust capacity to handle linguistic variations, which is essential for real-world applications where narrative inputs are often complex and subtly nuanced.

\section{Conclusion and Discussion}
This paper studied classification of Expressive Narrative Stories (ENS) within the mental health domain, highlighting the robust capabilities of advanced deep learning models. BERT(128) and MentalBERT(128) demonstrated superior performance compared to traditional machine learning methods, particularly in complex narrative analyses where explicit mental health-related keywords are sparse. Their resilience to textual manipulations, such as sentence shuffling, emphasizes their sophisticated understanding of narrative structures, crucial for applications in mental health assessments where narratives often exhibit intricate and nuanced language. This research underscores the necessity of  contextual understanding over mere keyword detection, advocating for a nuanced approach to mental health text analysis that probes into the subtle layers of narrative data.

The findings contribute to psycholinguistics and cognitive science by demonstrating that advanced models like BERT(128) can capture subtle linguistic cues that traditional models frequently overlook. The ability of BERT(128) to maintain high performance despite sentence manipulations supports theories that prioritize understanding the context and coherence of narratives in mental health diagnostics. This capability reflects a sophisticated cognitive processing attribute essential for models used in therapeutic settings, where discerning the logical flow and emotional undertones of patient narratives provides critical diagnostic insights.

Practically, implementing models such as BERT(128) and MentalBERT(128) in clinical environments could revolutionize the screening of mental health conditions through textual data analysis. Automated tools employing these models could analyze patient narratives submitted online, offering preliminary assessments that help prioritize cases or tailor therapeutic interventions. Moreover, these tools could facilitate ongoing monitoring of patient journals or digital communications as part of routine mental health care, providing clinicians with real-time insights into patients' mental states.

The study of ENS using these models offers a promising avenue for broad-scale mental health screening, particularly valuable in situations where healthcare professional availability is limited. By analyzing online texts—where individuals often express daily struggles and emotional experiences—these tools can preemptively identify individuals at risk who may not be aware of their need for mental health support. This proactive approach could significantly impact public health, especially for populations less likely to seek direct psychiatric help.

A notable observation from the study is MentalBERT's significant sensitivity to topic word manipulation, particularly when classifying between common ENS and depression-related ENS. This sensitivity highlights the complex linguistic characteristics that distinguish mental health narratives from more generic narrative content. Despite MentalBERT's design to handle psychiatric contexts, its performance variance in response to word manipulations underscores the challenges of deploying these models in real-world settings where explicit mental health terminology might be sparse.

Future research should focus on enhancing model robustness against linguistic variations and improving sensitivity to subtle narrative cues characteristic of different mental health conditions. Developing models that can effectively differentiate between types of ENS without heavily relying on explicit topic words could significantly advance mental health assessments conducted through text analysis.

This paper’s nuanced approach to narrative analysis enriches existing linguistic research, which often relies on simpler text analysis techniques such as keyword frequency and basic sentiment analysis. The deeper analytical capabilities of BERT(128) and MentalBERT(128) offer a robust framework for exploring the intricacies of mental health narratives, enhancing our understanding of how language is used in mental health contexts and paving the way for innovative diagnostic tools.


\section*{Acknowledgment}

We would like to acknowledge OpenAI for providing access to the ChatGPT model, which greatly assisted us to proofread and revise the text of the manuscript.

\bibliographystyle{IEEEtran}
\bibliography{bibli0612}

\end{document}

%% file: table_I.tex
\begin{table}[!htbp]
\caption{Sample Raw Text}
\label{tab:sample}
\begin{tabularx}{\columnwidth}{X}
\hline
\textbf {Sample Post Class \& Content} \\ \hline \hline
\textbf {GNS (Label: 0):} So I had some prime Denver steaks and I cut a couple up and braised them. They were amazing. So much so I went to the grocery store a week later and bought some grocery store choice chuck rib meat and braised that too. It was really good as well. I\textbackslash{}u2019ve only braised meat a handful of times so I don\textbackslash{}u2019t really know any standards. In my mind the prime cuts were better but my question is if it really makes a difference when slow cooking. Assuming I can afford it, am I wasting money braising some prime short ribs or is it really better in your experience? \\ \hline
\textbf {ENS-Common (Label: 0):} We\textbackslash{}u2019ve been in a relationship for almost 2 years now, and for the most part it\textbackslash{}u2019s been healthy \& happy.Unfortunately, before I met him I was in a string of bad relationships with men that cheated on me with their exes and were abusive. My most recent ex compared me to his exes, criticized my body, and more.My bf knows about this history, and I\textbackslash{}u2019ve expressed some discomfort that he still has photos of his ex, still has all his exes numbers in his phone, and also still has the numbers of girls he met from dating apps.I have no reason to distrust him and don\textbackslash{}u2019t want to be controlling. I know I need to work on this for myself. However, it hurts that he doesn\textbackslash{}u2019t seem to care about helping me feel more at ease when those small gestures would help. \\ \hline
\textbf {Depression-ENS (Label: 1):} I don't really know what to say here, just needed a place vent.Ive struggled with anxiety my whole life and depression for most of it. There's way more context to it than this but my family is made of well know scholars, both mom and dad are full professors at their universities, sister(27) is a honors master's student, aunt retired full professor, younger cousin(22) majoring in civil engineering and his brother(24), who's a mechanic engineer, just got in the best master program of our country. All of them went to either one of the two best federal university here in Brazil. I'm a 25yo communications student at a private university who's still one year away from graduating and decided not to do a masters for now. By all accounts my family see me as a deadbeat that has no future what so ever and makes zero effort to hide or constrain their opinion. Left my dads house a few years ago and had to return because life happen; The pandemic stretched my stay here from 1 year to 3 now and my relationship with my dad is at the limit. I don't have any idea on wtf I'm doing with my life and I'm starting to give up on everything.  \\ \hline
\end{tabularx}
\end{table}

%% file: table_II.tex
\begin{table}[!t]
\centering
\caption{Description of Training and Validation Datasets}
\label{tab:my-table}
\begin{tabular}{l|c|c|c}
\hline
\textbf{Attribute} & \textbf{GNS-Dep} & \textbf{ENS-Dep} & \textbf{Mix-Dep} \\ \hline
No. of Posts       & 2721           & 2186            & 3836           \\
Avg. Words/Post    & 99             & 145             & 122            \\
Max Word Count     & 1111           & 751             & 1111           \\
Ratio of Label     & 1.53           & 1.03            & 2.57           \\ \hline
\end{tabular}
\begin{minipage}{\columnwidth} 
\smallskip
Note: GNS-Dep = GNS-Depression, ENS-Dep = ENS-Depression, Mix-Dep = Mix-Depression.
\end{minipage}
\end{table}

%% file: table_III.tex
\begin{table}[!t]
\centering
\caption{List of Testing and Generalization Set}
\label{tab:my-table}
\begin{tabular}{lc}
\hline
\textbf{Dataset Name} & \textbf{Number of Posts} \\ \hline
GNS-Depression   & 3562            \\
ENS-Depression & 3070            \\
Mix-Depression       & 3976            \\
GNS-Anxiety      & 3565            \\
ENS-Anxiety    & 3070            \\
Mix-Anxiety          & 4220            \\
GNS-Suicide      & 3562            \\
ENS-Suicide    & 3070            \\
Mix-Suicide          & 6632            \\
GNS-Bipolar      & 3048            \\
ENS-Bipolar    & 3048            \\
Mix-Bipolar          & 3048            \\ 
\hline
\end{tabular}
\end{table}

%% file: table_IV.tex
\begin{table*}[!t]
\centering
\begin{threeparttable}
\caption{Comparative Accuracy of Various Models Across Different Depression Testing Sets}
\label{table:allmodels}
\begin{tabular}{lccc}
    \toprule
    Model & \multicolumn{3}{c}{Accuracy} \\ 
    \cmidrule(lr){2-4}
          & GNS-Depression Testing Set & ENS-Depression Testing Set & Mix-Depression Testing Set \\
    \midrule
    BERT(10) & 0.86 & 0.83 & 0.82 \\
    BERT(64) & \textbf{0.99} & 0.94 & 0.95 \\
    BERT(128) & \textbf{0.99} & \textbf{0.96} & \textbf{0.96} \\
    BERT(300) & \textbf{0.99} & \textbf{0.96} & 0.94 \\
    MentalBERT(10) & 0.86 & 0.80 & 0.81 \\
    MentalBERT(64) & 0.98 & 0.92 & 0.94 \\
    MentalBERT(128) & \textbf{0.99} & \textbf{0.96} & \textbf{0.96} \\
    MentalBERT(300) & \textbf{0.99} & 0.95 & 0.95 \\
    NB(TDIDF: 5000) & \textbf{0.99} & 0.94 & 0.90 \\
    SVM(TDIDF: 5000) & \textbf{0.99} & 0.95 & 0.93 \\
    LR(TDIDF: 5000) & 0.97 & 0.94 & 0.90 \\
    NB(unigram) & \textbf{0.99} & 0.93 & 0.93 \\
    SVM(unigram) & 0.97 & 0.93 & 0.92 \\
    \bottomrule
   \end{tabular}
    \begin{tablenotes}
\footnotesize 
\item[*] Bold numbers indicate the highest accuracy for each test set.
\item[*] Parameters in parentheses are model configurations: 'BERT' and 'MentalBERT' numbers specify input sequence lengths. 'TDIDF: 5000' and 'Unigram' denote the type of text processing used.
\end{tablenotes}
\end{threeparttable}
\end{table*}

%% file: table_V.tex
\begin{table}[!t]
\centering
\caption{Example of Topic Word Manipulations}
\label{tab:my-table}
\begin{tabular}{p{3.4cm}|p{4.6cm}} 
\hline
\textbf{Words Process}                & \textbf{Sentence}                                        \\ \hline
Raw                                   & i have very good relationship with my friend. \\ \hline
Words Removing                        & i have very good with my.                     \\ \hline
Words Replacing to “nothing” & i have very good nothing with my nothing.    \\ \hline
\end{tabular}
\end{table}

%% file: table_VI.tex
\begin{sidewaystable*}[t]  
\centering
\caption{Performance Change and T-test Result of Word Manipulations across Different Models and Different Testing Sets.}
\label{table:performance_results}
\begin{threeparttable}
\small  
\setlength\tabcolsep{1pt} 
\begin{tabular}{llcccccccccccc}
\toprule
\multicolumn{1}{l}{\multirow{2}{*}{Model}} & \multicolumn{1}{l}{\multirow{2}{*}{Word Manipulation}} & \multicolumn{4}{c}{GNS-Depression Testing Set} & \multicolumn{4}{c}{ENS-Depression Testing Set} & \multicolumn{4}{c}{Mix-Depression Testing Set} \\ 
\cmidrule(lr){3-6} \cmidrule(lr){7-10} \cmidrule(lr){11-14}
& & Accuracy(\%) & AccDiff(\%) & t-statistics & P-value & Accuracy(\%) & AccDiff(\%) & t-statistics & P-value & Accuracy(\%) & AccDiff(\%) & t-statistics & P-value \\
\midrule
\multirow{3}{*}{BERT(128)} & Raw & 99.3 & - & - & - & 95.9 & - & - & - & 95.9 & - & - & - \\
& Word Removing & 98.8 & -0.5 & -0.79 & 0.43 & 95.3 & -0.6 & -0.20 & 0.84 & 94.6 & -1.3 & 2.14 & \textbf{0.03*} \\
& Word Replacing & 98.4 & -0.9 & -2.11 & \textbf{0.04*} & 95.3 & -0.6 & -0.55 & 0.58 & 94.4 & -1.5 & 1.43 & 0.67 \\
\midrule
\multirow{3}{*}{MentalBERT(128)} & Raw & 99.4 & - & - & - & 96.0 & - & - & - & 96.2 & - & - & - \\
& Word Removing & 98.9 & -0.5 & -1.40 & 0.16 & 95.4 & -0.6 & 3.14 & \textbf{\textless{}0.01**} & 95.1 & -1.1 & 3.75 & \textbf{\textless{}0.01**} \\
& Word Replacing & 98.1 & -1.3 & -3.99 & \textbf{\textless{}0.01**} & 95.3 & -0.7 & 1.15 & 0.25 & 94.3 & -1.9 & 5.38 & \textbf{\textless{}0.01**} \\
\midrule
\multirow{3}{*}{NB(TDIDF: 5000)} & Raw & 98.6 & - & - & - & 93.8 & - & - & - & 90.2 & - & - & - \\
& Word Removing & 98.2 & -0.4 & -0.22 & 0.83 & 93.3 & -0.5 & 1.22 & 0.22 & 88.3 & -1.9 & 8.14 & \textbf{\textless{}0.01**} \\
& Word Replacing & 98.0 & -0.6 & -4.33 & \textbf{\textless{}0.01**} & 93.0 & -0.8 & -2.47 & \textbf{\textless{}0.01**} & 88.1 & -2.1 & 7.80 & \textbf{\textless{}0.01**} \\
\midrule
\multirow{3}{*}{SVM(TDIDF: 5000)} & Raw & 98.5 & - & - & - & 94.7 & - & - & - & 93.4 & - & - & - \\
& Word Removing & 97.6 & -0.9 & 0.34 & 0.73 & 94.0 & -0.7 & 2.14 & \textbf{0.03*} & 91.2 & -2.2 & 6.30 & \textbf{\textless{}0.01**} \\
& Word Replacing & 97.0 & -1.5 & -4.10 & \textbf{\textless{}0.01**} & 93.7 & -1.0 & -0.87 & 0.39 & 91.6 & -1.8 & 3.09 & \textbf{\textless{}0.01**} \\
\midrule
\multirow{3}{*}{LR(TDIDF: 5000)} & Raw & 96.5 & - & - & - & 93.7 & - & - & - & 90.3 & - & - & - \\
& Word Removing & 95.0 & -1.5 & 4.55 & \textbf{\textless{}0.01**} & 93.9 & 0.2 & 2.12 & \textbf{0.03*} & 87.0 & -3.3 & 11.60 & \textbf{\textless{}0.01**} \\
& Word Replacing & 94.5 & -2.0 & 3.27 & \textbf{\textless{}0.01**} & 92.6 & -1.1 & -0.15 & 0.88 & 86.5 & -0.5 & 10.43 & \textbf{\textless{}0.01**} \\
\midrule
\multirow{3}{*}{NB(unigram)} & Raw & 99.3 & - & - & - & 93.3 & - & - & - & 93.1 & - & - & - \\
& Word Removing & 98.7 & -0.6 & -1.15 & 0.25 & 91.4 & -1.9 & 2.54 & \textbf{\textless{}0.01**} & 92.3 & -0.8 & 3.00 & \textbf{\textless{}0.01**} \\
& Word Replacing & 98.1 & -1.2 & -3.62 & \textbf{\textless{}0.01**} & 90.3 & -3.0 & 0.61 & 0.54 & 91.0 & -2.1 & 1.84 & \textbf{0.07*} \\
\midrule
\multirow{3}{*}{SVM(unigram)} & Raw & 97.2 & - & - & - & 93.1 & - & - & - & 91.9 & - & - & - \\
& Word Removing & 99.5 & 2.3 & 2.31 & \textbf{0.02*} & 91.6 & -1.5 & 2.29 & \textbf{0.02*} & 88.8 & -3.1 & 8.94 & \textbf{\textless{}0.01**} \\
& Word Replacing & 93.7 & -3.5 & 3.41 & \textbf{\textless{}0.01**} & 90.1 & -3.0 & -1.49 & 0.14 & 86.6 & -5.3 & 9.02 & \textbf{\textless{}0.01**} \\
\bottomrule
\end{tabular}
\begin{tablenotes}
\small
\item Note: AccDiff indicates the difference of accuracy between raw data and word manipulation. Bold numbers indicate the significant differences between raw data and the manipulated data. *P-value\textless{}=0.05, **P-value\textless{}0.01.
\end{tablenotes}
\end{threeparttable}
\end{sidewaystable*}

%% file: table_VII.tex
\begin{table*}[!t]
\centering
\begin{threeparttable}
\caption{Performance Change and T-test Result of Sentence Manipulations across Different Models and Different Testing Sets.}
\small
\label{table:performance_change}
\begin{tabular}{@{}l|l|l|cccc@{}}
\toprule
Model & Testing Set & Sentence Manipulation & Accuracy(\%) & AccDiff(\%) & T Score & P-value \\ 
\midrule
\multirow{9}{*}{BERT(128)-GNS} & \multirow{3}{*}{GNS-Depression} & Raw & 99.3 & - & - & - \\
 &  & Cross-Post Shuffling & 100.0 & 0.7 & -0.46 & 0.65 \\
 &  & Within-Post Shuffling & 99.4 & 0.1 & 0.75 & 0.45 \\ 
\cmidrule(l){2-7} 
 & \multirow{3}{*}{GNS-Anxiety} & Raw & 99.4 & - & - & - \\
 &  & Cross-Post Shuffling & 100.0 & 0.6 & -0.47 & 0.64 \\
 &  & Within-Post Shuffling & 99.1 & -0.3 & 0.18 & 0.86 \\ 
\cmidrule(l){2-7} 
 & \multirow{3}{*}{GNS-Suicide} & Raw & 99.4 & - & - & - \\
 &  & Cross-Post Shuffling & 99.9 & 0.5 & -0.43 & 0.67 \\
 &  & Within-Post Shuffling & 99.4 & 0.0 & 0.04 & 0.97 \\ 
\cmidrule(l){2-7} 
 & \multirow{3}{*}{GNS-Bipolar} & Raw & 99.5 & - & - & - \\
 &  & Cross-Post Shuffling & 100.0 & 0.5 & -0.58 & 0.56 \\
 &  & Within-Post Shuffling & 99.5 & 0.0 & 0.12 & 0.90 \\ 
\midrule
\multirow{9}{*}{BERT(128)-ENS} & \multirow{3}{*}{ENS-Depression} & Raw & 95.9 & - & - & - \\
 &  & Cross-Post Shuffling & 94.8 & -1.1 & -1.79 & \textbf{\textless{}0.01**} \\
 &  & Within-Post Shuffling & 94.4 & -1.5 & -3.00 & \textbf{0.05*} \\ 
\cmidrule(l){2-7} 
 & \multirow{3}{*}{ENS-Anxiety} & Raw & 95.7 & - & - & - \\
 &  & Cross-Post Shuffling & 95.2 & -0.5 & -2.01 & \textbf{\textless{}0.01**} \\
 &  & Within-Post Shuffling & 94.6 & -1.1 & -3.33 & \textbf{0.04*} \\ 
\cmidrule(l){2-7} 
 & \multirow{3}{*}{ENS-Suicide} & Raw & 96.2 & - & - & - \\
 &  & Cross-Post Shuffling & 94.8 & -1.4 & -2.19 & \textbf{\textless{}0.01**} \\
 &  & Within-Post Shuffling & 95.1 & -1.1 & -2.68 & \textbf{0.03*} \\ 
\cmidrule(l){2-7} 
 & \multirow{3}{*}{ENS-Bipolar} & Raw & 96.0 & - & - & - \\
 &  & Cross-Post Shuffling & 95.3 & -0.7 & -2.08 & \textbf{\textless{}0.01**} \\
 &  & Within-Post Shuffling & 94.1 & -1.9 & -2.93 & \textbf{0.04*} \\ 
\midrule

\multirow{12}{*}{MentalBERT(128)-ENS} & \multirow{3}{*}{ENS-Depression} & Raw & 96.0 & - & - & - \\
 &  & Cross-Post Shuffling & 96.7 & 0.7 & -1.29 & 0.20 \\
 &  & Within-Post Shuffling & 94.6 & -1.4 & -2.92 & \textbf{\textless{}0.01**} \\ 
\cmidrule(l){2-7} 
 & \multirow{3}{*}{ENS-Anxiety} & Raw & 95.9 & - & - & - \\
 &  & Cross-Post Shuffling & 97.2 & 1.3 & -1.73 & \textbf{0.08*} \\
 &  & Within-Post Shuffling & 95.4 & -0.5 & -5.79 & \textbf{\textless{}0.01**} \\ 
\cmidrule(l){2-7} 
 & \multirow{3}{*}{ENS-Suicide} & Raw & 96.0 & - & - & - \\
 &  & Cross-Post Shuffling & 96.7 & 0.7 & -0.91 & 0.36 \\
 &  & Within-Post Shuffling & 95.0 & -1.0 & -4.28 & \textbf{\textless{}0.01**} \\ 
\cmidrule(l){2-7} 
 & \multirow{3}{*}{ENS-Bipolar} & Raw & 96.7 & - & - & - \\
 &  & Cross-Post Shuffling & 97.2 & 0.5 & -1.57 & 0.12 \\
 &  & Within-Post Shuffling & 95.6 & -1.1 & -4.87 & \textbf{\textless{}0.01**} \\ 
\bottomrule
\end{tabular}
\smallskip
\begin{tablenotes}
            \small
            \item Note: AccDiff indicates the difference of accuracy between raw data and sentence manipulation. Bold numbers indicate the significant differences between raw data and the manipulated data. Note: *P-value\textless{}=0.05, **P-value\textless{}0.01
\end{tablenotes}
\end{threeparttable}
\end{table*}